\pdfoutput=1

\documentclass[11pt]{article}

\usepackage{acl}

\usepackage{times}
\usepackage{latexsym}

\usepackage[T1]{fontenc}

\usepackage[utf8]{inputenc}

\usepackage{microtype}

\usepackage{inconsolata}

\usepackage{amsmath}
\usepackage{amssymb}
\usepackage{xspace}
\usepackage{url}
\usepackage{booktabs}
\usepackage{multirow}
\usepackage{multicol}
\usepackage{makecell}
\usepackage{xcolor}
\usepackage{graphicx}
\usepackage{color}
\usepackage{etoolbox}

\newcommand{\name}{ComSpeech\xspace}

\title{Can We Achieve High-quality Direct Speech-to-Speech Translation without Parallel Speech Data?}

\author{
    Qingkai Fang$^{1,3}$,
    Shaolei Zhang$^{1,3}$,
    Zhengrui Ma$^{1,3}$,
    Min Zhang$^{4}$,
    Yang Feng$^{1,2,3}$\thanks{Corresponding author: Yang Feng.} \\
    \textsuperscript{\rm1}Key Laboratory of Intelligent Information Processing \\ Institute of Computing Technology, Chinese Academy of Sciences (ICT/CAS) \\
    \textsuperscript{\rm2}Key Laboratory of AI Safety, Chinese Academy of Sciences \\
    \textsuperscript{\rm3}University of Chinese Academy of Sciences, Beijing, China \\
    \textsuperscript{\rm4}School of Future Science and Engineering, Soochow University \\
    {$\;\:$\texttt{\{\href{mailto:fangqingkai21b@ict.ac.cn}{fangqingkai21b},\href{mailto:fengyang@ict.ac.cn}{fengyang}\}@ict.ac.cn}}
    {$\;\:$\texttt{\href{mailto:zhangminmt@hotmail.com}{zhangminmt}@hotmail.com}}
}

\begin{document}
\maketitle
\begin{abstract}
\makeatletter
\patchcmd{\@makefntext}{\hspace{1em}}{}{}{}
\makeatother

Recently proposed two-pass direct speech-to-speech translation (S2ST) models decompose the task into speech-to-text translation (S2TT) and text-to-speech (TTS) within an end-to-end model, yielding promising results. However, the training of these models still relies on \textit{parallel speech data}, which is extremely challenging to collect. 
In contrast, S2TT and TTS have accumulated a large amount of data and pretrained models, which have not been fully utilized in the development of S2ST models.
Inspired by this, in this paper, we first introduce a composite S2ST model named \name, which can seamlessly integrate any pretrained S2TT and TTS models into a direct S2ST model. 
Furthermore, to eliminate the reliance on parallel speech data, we propose a novel training method \name-ZS that solely utilizes S2TT and TTS data. It aligns representations in the latent space through contrastive learning, enabling the speech synthesis capability learned from the TTS data to generalize to S2ST in a zero-shot manner.
Experimental results on the CVSS dataset show that when the parallel speech data is available, \name surpasses previous two-pass models like UnitY and Translatotron 2 in both translation quality and decoding speed. When there is no parallel speech data, \name-ZS lags behind \name by only 0.7 ASR-BLEU and outperforms the cascaded models.\footnote{Project Page: \hypersetup{urlcolor=magenta}\href{https://ictnlp.github.io/ComSpeech-Site/}{https://ictnlp.github.io/ComSpeech-Site/}}
\end{abstract}

\section{Introduction}
Direct speech-to-speech translation (S2ST) refers to the process of translating source language speech directly into target language speech using an end-to-end model~\citep{translatotron}. In contrast to traditional cascaded S2ST~\citep{Wahlster2000, 1597243}, which comprises a pipeline of automatic speech recognition (ASR), machine translation (MT), and text-to-speech (TTS), direct S2ST has the advantage of reducing errors accumulation between modules and being easier to deploy~\citep{s2ut}. Therefore, it has attracted considerable interest from researchers in recent years~\citep{hokkien, huang2023chch, fang-etal-2023-daspeech}.

S2ST needs to learn the mapping between speech in different languages. However, due to the complex distribution of speech data, learning such mapping presents significant challenges. To address this, \citet{translatotron2} introduces the two-pass direct S2ST model, which decomposes S2ST into the subtasks of speech-to-text translation (S2TT) and TTS within an end-to-end model. It begins by generating the target text using one decoder, followed by generating the target speech using another decoder. This modeling approach reduces learning complexity while retaining the benefits of end-to-end modeling, surpassing the performance of cascaded S2ST~\citep{inaguma-etal-2023-unity}.


Despite the success of two-pass S2ST models, they still face a significant challenge: the training of these models still relies on \textit{parallel speech data}, which is extremely difficult to collect. In fact, current S2ST datasets are mostly derived from existing S2TT datasets by converting the target text into speech using TTS models, which is very time-consuming and becomes impractical when dealing with large-scale S2TT datasets. In contrast, S2TT and TTS have accumulated numerous datasets and pretrained models through past developments. A natural question is how to utilize those pretrained models when constructing an S2ST model, thereby maximizing the utilization of existing resources. 
Intuitively, we can feed the representations output by the S2TT model into the TTS encoder, thereby combining the pretrained S2TT and TTS models into a two-pass S2ST model.
However, one major challenge is that S2TT and TTS models typically use different vocabularies for the target language. For example, S2TT models often use subword vocabularies, while TTS models often use phoneme or character vocabularies. 
Inconsistent vocabularies lead to different segmentations of the same target text. Consequently, directly combining them into an S2ST model results in the TTS model being unable to process the representations output by the S2TT model.
Even if we can utilize S2TT and TTS models with the same vocabulary, it is still necessary to finetune the overall model with parallel speech data to adapt it to the S2ST task. 

To address the aforementioned challenges, in this paper, we first propose a composite speech-to-speech translation model, named \name. 
By introducing a vocabulary adaptor based on connectionist temporal classification~\citep[CTC;][]{ctc}, it facilitates the conversion of representation sequences between any vocabularies, thus enabling the integration of any pretrained S2TT and TTS models into an S2ST model.
Furthermore, to eliminate the dependence on parallel speech data during training, we propose a novel training strategy \name-ZS that uses only S2TT and TTS data. 
It employs contrastive learning to align representations in the latent space, allowing the speech synthesis capabilities acquired from the TTS data to generalize to S2ST.
Experimental results on the CVSS dataset demonstrate that \name surpasses previous two-pass models in both translation quality and decoding speed. In the zero-shot learning scenario, the translation quality of \name-ZS is only 0.7 ASR-BLEU lower than \name and surpasses cascaded systems. 

\section{Background: Two-pass S2ST}

The S2ST dataset usually contains three-way parallel data $\mathcal{D}_{\text{S2ST}} = \{(S, Y, T)\}$, where $S$ denotes the source speech, $Y$ is the target text, and $T$ is the target speech. 
Due to the challenges of directly generating target speech, recent developments have introduced two-pass S2ST models, which enhance translation quality by incorporating the target text into the generation process~\citep{translatotron2, inaguma-etal-2023-unity}.
Formally, a two-pass model typically consists of four sub-modules: the source speech encoder $\mathcal{F}_{\text{enc}}$, the target text decoder $\mathcal{F}_{\text{dec}}$, the text-to-speech encoder $\mathcal{G}_{\text{enc}}$, and the target speech decoder $\mathcal{G}_{\text{dec}}$. Firstly, $\mathcal{F}_{\text{enc}}$ and $\mathcal{F}_{\text{dec}}$ perform the S2TT task, trained with the cross-entropy loss:
\begin{equation}
\mathcal{L}_{\text{S2TT}} = -\sum_{i=1}^{|Y|}\log P(y_i | \mathcal{F}_{\text{dec}}(\mathcal{F}_{\text{enc}}(S), Y_{<i})).
\end{equation}
Subsequently, $\mathcal{G}_{\text{enc}}$ takes the representation from the last layer of $\mathcal{F}_{\text{dec}}$ as its input, and then $\mathcal{G}_{\text{dec}}$ predicts the target speech. The training objective is to minimize the distance between the predicted speech and the ground truth speech:
\begin{equation}
    \mathcal{L}_{\text{S2ST}}= d(T, \mathcal{G}_{\text{dec}} \circ \mathcal{G}_{\text{enc}} \circ \mathcal{F}_{\text{dec}}(\mathcal{F}_{\text{enc}}(S), Y_{<i})),
\end{equation}
where $d(\cdot, \cdot)$ denotes the distance function, with its specific form determined by the representation type of the target speech (e.g., discrete units or mel-spectrograms). In summary, the two-pass S2ST model is trained on the parallel speech dataset $\mathcal{D}_{\text{S2ST}}$ with the following objective:
\begin{equation}
    \mathcal{L}_{\text{Two-Pass}} = \mathcal{L}_{\text{S2TT}} + \gamma\cdot \mathcal{L}_{\text{S2ST}},
\end{equation}
where $\gamma$ denotes the weight of the second term.
However, \textbf{current two-pass S2ST models still have two main issues}: (1) First, they implicitly assumes that $\mathcal{F}_{\text{dec}}$ and $\mathcal{G}_{\text{enc}}$ share the same target text vocabulary, making it challenging to simultaneously employ an existing S2TT model $\mathcal{F}$ as $\mathcal{F}_{\text{enc}}$ and $\mathcal{F}_{\text{dec}}$, while utilizing an existing TTS model $\mathcal{G}$ as $\mathcal{G}_{\text{enc}}$ and $\mathcal{G}_{\text{dec}}$. (2) Second, their training requires parallel speech data, which is extremely difficult to collect.
To address these issues, we first propose a composite S2ST model built upon existing S2TT and TTS models (Section~\ref{sec:model}). Next, to eliminate the reliance on parallel speech data, we introduce a novel training method that uses only S2TT and TTS data to achieve zero-shot S2ST (Section~\ref{sec:training}).



\section{Proposed Model: \name}
\label{sec:model}

\begin{figure}[t]
    \centering
    \includegraphics[width=\linewidth]{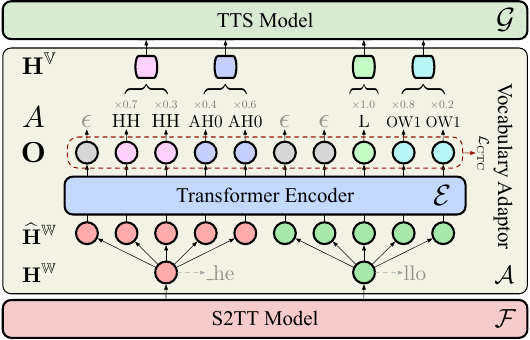}
    \caption{Model architecture of our proposed \name. It includes an S2TT model $\mathcal{F}$, a TTS model $\mathcal{G}$, and a vocabulary adaptor $\mathcal{A}$ to connect $F$ and $G$.}
    \label{fig:merge}
\end{figure}

In this section, we introduce a more general two-pass S2ST model architecture: Composite Speech-to-Speech Translation (\name) model. \name comprises three modules: $\mathcal{F}$, $\mathcal{A}$, and $\mathcal{G}$, where $\mathcal{F}$ denotes arbitrary S2TT model, $\mathcal{G}$ denotes arbitrary TTS model for the target language, and $\mathcal{A}$ represents the \textit{vocabulary adaptor} inserted between $\mathcal{F}$ and $\mathcal{G}$. The inclusion of the vocabulary adaptor $\mathcal{A}$ is necessitated by the fact that S2TT and TTS models typically employ different vocabularies for the target language. It can facilitate the conversion of target text representation sequences between different vocabularies. Figure \ref{fig:merge} illustrates the model architecture of \name.

Specifically, we use $\mathbb{W}$ and $\mathbb{V}$ to denote the vocabularies of the S2TT model and the TTS model, respectively. The tokenized sequences of the target text $Y$ under these two vocabularies are represented as $Y^\mathbb{W} = (y^\mathbb{W}_1, ..., y^\mathbb{W}_N)$ and $Y^\mathbb{V} = (y^\mathbb{V}_1, ..., y^\mathbb{V}_M)$, respectively. Next, we will introduce the detailed implementations of each module in \name.
\paragraph{S2TT Model} We use the \texttt{s2t\_conformer} model in \textit{fairseq S2T}\footnote{\url{https://github.com/facebookresearch/fairseq/tree/main/examples/speech_to_text}}~\citep{wang2020fairseqs2t} as the S2TT model $\mathcal{F}$. Specifically, it comprises a Conformer-based speech encoder~\citep{conformer} and a Transformer-based text decoder~\citep{transformer}. The decoder adopts a subword vocabulary $\mathbb{W}$ for the target language. Formally, the decoder output hidden states of the S2TT model can be represented as $\mathbf{H}^\mathbb{W}=(\mathbf{h}^\mathbb{W}_1, ..., \mathbf{h}^\mathbb{W}_N)$, where $\mathbf{h}^\mathbb{W}_i = \mathcal{F}(S, Y^\mathbb{W}_{<i})$. The S2TT model is trained by minimizing the cross-entropy loss:
\begin{equation}
    \mathcal{L}_\text{S2TT} = -\sum_{i=1}^{N}\log P(y^\mathbb{W}_i | \mathcal{F}(S, Y^\mathbb{W}_{<i})).
\end{equation}

\paragraph{TTS Model} We use FastSpeech 2~\citep{ren2021fastspeech2} as the TTS model $\mathcal{G}$.
Specifically, it consists of an encoder, a variance adapter, and a decoder. 
FastSpeech 2 adopts a phoneme vocabulary $\mathbb{V}$ for the target language. For the independent TTS model, the encoder takes the target text embedding $\textsc{Emb}(Y^\mathbb{V})$ as input, where $\textsc{Emb}(\cdot)$ denotes the phoneme embedding layer. 
The variance adaptor and decoder predict variance information and target mel-spectrogram, respectively. The training objective of FastSpeech 2 can be written as follows:
\begin{equation}
    \mathcal{L}_\text{TTS} = \mathcal{L}_\text{L1} + \mathcal{L}_\text{dur} + \mathcal{L}_\text{pitch} + \mathcal{L}_\text{energy},
\end{equation}
where $\mathcal{L}_\text{L1}$ calculates the L1 distance between the predicted and ground truth mel-spectrograms, $\mathcal{L}_\text{dur}$, $\mathcal{L}_\text{pitch}$, and $\mathcal{L}_\text{energy}$ denote the mean square error (MSE) loss between ground truth and predictions for duration, pitch, and energy, respectively.

When we combine $\mathcal{F}$ and $\mathcal{G}$ into a two-pass S2ST model, the TTS encoder $\mathcal{G}_\text{enc}$ no longer takes $\textsc{Emb}(Y^\mathbb{V})$ as input but instead accepts the representations $\mathbf{H}^{\mathbb{W}}$ from the S2TT model. However, $\mathbf{H}^{\mathbb{W}}$ corresponds to $Y^{\mathbb{W}}$ and cannot be directly understood by the TTS encoder due to the vocabulary mismatch. Therefore, we introduce a vocabulary adaptor to convert the representation sequence corresponding to $Y^{\mathbb{W}}$ into that corresponding to $Y^{\mathbb{V}}$.

\begin{figure*}[t]
    \centering
    \includegraphics[width=\textwidth]{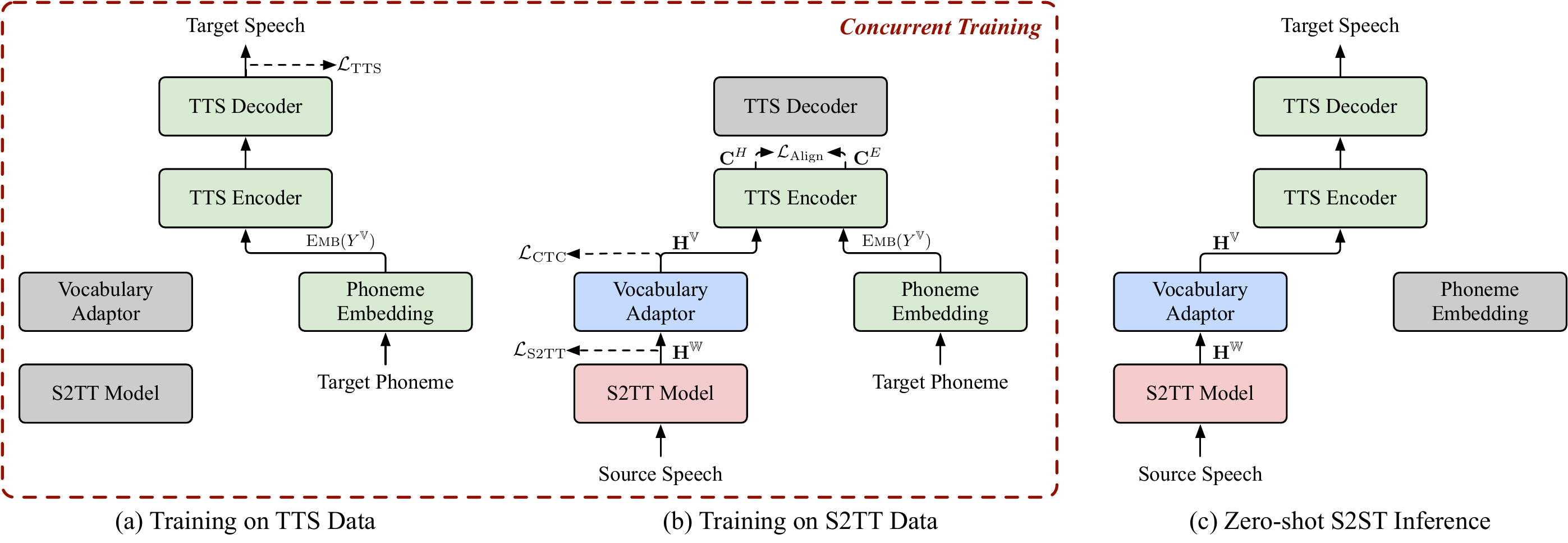}
    \caption{Illustration of training and inference process in the zero-shot learning scenario. Solid lines represent data flow, while dashed lines represent loss calculation. The gray modules do not participate in the computation.}
    \label{fig:training}
\end{figure*}

\paragraph{Vocabulary Adaptor} 
To facilitate variable-length sequence conversion between different vocabularies ($\mathbb{W}\rightarrow\mathbb{V}$), we introduce a vocabulary adaptor $\mathcal{A}$ based on connectionist temporal classification ~\citep[CTC;][]{ctc}. 
Specifically, we first upsample each hidden state in $\mathbf{H}^\mathbb{W}$ by a factor of $\lambda$, resulting in an upsampled hidden state sequence $\widehat{\mathbf{H}}^\mathbb{W}=(\widehat{\mathbf{h}}^\mathbb{W}_1, ..., \widehat{\mathbf{h}}^\mathbb{W}_{\lambda\cdot N})$, where $\widehat{\mathbf{h}}^\mathbb{W}_i = \mathbf{h}^\mathbb{W}_{\lfloor i/\lambda\rfloor}$. Next, we add positional encoding to $\widehat{\mathbf{H}}^\mathbb{W}$ and then use an $L$-layer Transformer encoder $\mathcal{E}$ for encoding, obtaining the output hidden state sequence $\mathbf{O} = (\mathbf{o}_1, \ldots, \mathbf{o}_{\lambda \cdot N})$:
\begin{equation}
    \mathbf{O} = \mathcal{E} (\widehat{\mathbf{H}}^\mathbb{W} + \textsc{Pos} (\widehat{\mathbf{H}}^\mathbb{W})),
\end{equation}
where $\textsc{Pos}(\cdot)$ denotes the sinusoid positional encoding~\citep{transformer}. To achieve cross-vocabulary conversion, we aim to align $\mathbf{O}$ with the target text $Y^\mathbb{V}$ tokenized by the TTS vocabulary $\mathbb{V}$. Thus, we use CTC to model the alignment between the two sequences. Specifically, CTC extends the output space $\mathbb{V}$ with a special blank token $\epsilon$, and maps each element in $\mathbf{O}$ into the output space:
\begin{equation}
    P(a_i | \mathbf{O}) = \text{softmax} (\mathbf{W}\mathbf{o}_i + \mathbf{b})[a_i]\ \forall a_i\in \mathbb{V}\cup\{\epsilon\},
\end{equation}
where $\mathbf{W}\in\mathbb{R}^{(|\mathbb{V}| + 1)\times d}$ and $\mathbf{b}\in\mathbb{R}^{|\mathbb{V}| + 1}$ are the weights and biases of the linear layer, and $A=(a_1, ..., a_{\lambda\cdot N})$ is referred to as the \textit{alignment}. CTC introduces a collapsing function $\beta(A)$ that initially merges all consecutively repeated tokens in $A$ and then removes all blank tokens $\epsilon$. For example: $\beta(\texttt{aab}\epsilon\epsilon\texttt{bbc}) = \texttt{abbc}$. During training, CTC marginalizes out all possible alignments as follows:
\begin{equation}
\begin{aligned}
    \mathcal{L}_\text{CTC} &= -\log P(Y^\mathbb{V} | \mathbf{O}) \\
    &= -\log \sum_{A\in\beta^{-1}(Y^\mathbb{V})}  P(A|\mathbf{O}) \\
    &= -\log \sum_{A\in\beta^{-1}(Y^\mathbb{V})} \prod_{i=1}^{\lambda\cdot N} P(a_i|\mathbf{O}),
\end{aligned}
\end{equation}
where $\beta^{-1}(Y^\mathbb{V})$ denotes all possible alignments of length $\lambda\cdot N$ that can be collapsed to $Y^\mathbb{V}$.

Through CTC, we can learn the alignment between the output representation sequence $\mathbf{O}$ and the target text sequence $Y^{\mathbb{V}}$. 
However, the length of $\mathbf{O}$ is greater than or equal to $Y^{\mathbb{V}}$ ($\lambda\cdot N\geq M$) due to the presence of repeated and blank tokens in the alignment $A$. 
During training, it is essential to obtain the representation sequence exactly corresponding to the ground truth target $Y^\mathbb{V}$, which enables joint training with the subsequent TTS model. 
Therefore, we compress $\mathbf{O}$ to length $M$ following these steps: (1) Firstly, we find the most probable alignment $A^* = \mathop{\arg\max}_{A} P(A|\mathbf{O}, Y^\mathbb{V})$ via Viterbi algorithm~\citep{viterbi}, which is often referred to as \textit{forced alignment}. (2) Secondly, each continuous repetition of non-blank tokens in $A^*$ is divided into a segment. In this way, each token $y^\mathbb{V}_i$ in $Y^\mathbb{V}$ corresponds to a segment $[a^*_{l_i}, ..., a^*_{r_i}]$, where $a^*_j = y^\mathbb{V}_i, \forall l_i\leq j\leq r_i$. (3) Finally, the representations within each segment are merged into a vector according to the prediction confidence~\citep{liu2020_bridging, gaido-etal-2021-ctc}:
\begin{equation}
    \mathbf{h}^\mathbb{V}_i = \sum_{j=l_i}^{r_i} p_j\cdot \mathbf{o}_j,
\end{equation}
\begin{equation}
    p_j = \frac{\exp(P(a^*_j | \mathbf{O}))}{\sum_{k=l_i}^{r_i} \exp(P(a^*_{k} | \mathbf{O}))}.
\end{equation}
The representations corresponding to blank tokens are discarded. Finally, we obtain the representation sequence $\mathbf{H}^\mathbb{V} = (\mathbf{h}^\mathbb{V}_1, ..., \mathbf{h}^\mathbb{V}_M)$ corresponding to $Y^\mathbb{V}$, which is fed into the TTS encoder for subsequent speech synthesis. During inference, we select the most probable path $A^* = (a^*_1, ..., a^*_{\lambda\cdot N})$ with argmax decoding: $a^*_i = \mathop{\arg\max}_{a_i} P(a_i | \mathbf{O})$, and then merge representations in the same manner as during training.


\paragraph{Training} \name can be trained using S2TT, TTS, and S2ST data. Specifically, we can utilize S2TT data to pretrain the S2TT model $\mathcal{F}$ and TTS data to pretrain the TTS model $\mathcal{G}$. Subsequently, the entire model is finetuned using S2ST data. The training objective during finetuning is as follows:
\begin{equation}
    \mathcal{L}_\text{\name} = \mathcal{L}_\text{S2TT} + \mathcal{L}_\text{CTC} + \mathcal{L}_\text{TTS}.
\end{equation}
Moreover, as the vocabulary adaptor can establish connections between arbitrary S2TT and TTS models, theoretically, we can leverage any pretrained S2TT and TTS models for S2ST finetuning.

\section{\name-ZS: Training \name without Parallel Speech Data}
\label{sec:training}

As mentioned earlier, collecting S2ST data is extremely challenging. In this section, we explore the possibility of training \name exclusively on S2TT and TTS data, achieving \textit{zero-shot} S2ST without the need for parallel speech data. The idea is to train distinct modules using S2TT and TTS data concurrently, while achieving representation alignment in the output space of the TTS encoder. 
This enables the speech synthesis capabilities learned from the TTS data to generalize to S2ST during inference.
Figure \ref{fig:training} illustrates the process of training and inference.

Specifically, we use $\mathcal{D}_\text{S2TT}=\{(S, Y)\}$ and $\mathcal{D}_\text{TTS}=\{(Y, T)\}$ to denote the S2TT and TTS datasets, respectively, and train the model concurrently using both datasets. For TTS data batches, we train the TTS model $\mathcal{G}$ with $\mathcal{L}_\text{TTS}$, as illustrated in Figure \ref{fig:training}(a). For S2TT data batches, firstly, we train the S2TT model $\mathcal{F}$ and vocabulary adaptor $\mathcal{A}$ with $\mathcal{L}_\text{S2TT}$ and $\mathcal{L}_\text{CTC}$. Secondly, since the TTS encoder receives the target text embedding $\textsc{Emb}(Y^\mathbb{V})$ during TTS training but vocabulary adaptor's output $\mathbf{H}^\mathbb{V}$ during inference, we aim for the output representations of the TTS encoder with these two inputs to be identical, thereby achieving zero-shot generalization.
Therefore, we introduce a representation alignment loss after the TTS encoder during S2TT training, as illustrated in Figure \ref{fig:training}(b).

\paragraph{Representation Alignment}
Formally, we use $\mathbf{C}^H = (\mathbf{c}^H_1, ..., \mathbf{c}^H_M)$ and $\mathbf{C}^E = (\mathbf{c}^E_1, ..., \mathbf{c}^E_M)$ to denote the output representations of TTS encoder with these two types of input:
\begin{equation}
    \mathbf{C}^H = \mathcal{G}_\text{enc} (\mathbf{H}^\mathbb{V}), \ \ \mathbf{C}^E = \mathcal{G}_\text{enc} (\textsc{Emb}(Y^\mathbb{V})).
\end{equation}
Firstly, we align representations with the MSE loss:
\begin{equation}
    \mathcal{L}_\text{MSE} = \sum_{i=1}^M \| \mathbf{c}^H_i - \mathbf{c}^E_i \|^2_2.
\end{equation}
Additionally, we incorporate a contrastive learning objective to align representations. This objective maximizes the similarity between representations from the same input pairs and minimizes the similarity between representations from different input pairs. The contrastive loss is formulated as:
\begin{equation}
\begin{aligned}
    \mathcal{L}_\text{CTR} = &-\frac{1}{2}\sum_{i=1}^M \log \frac{\exp(s(\mathbf{c}^H_i, \mathbf{c}^E_i) / \tau)}{\sum_{j=1}^M \exp (s(\mathbf{c}^H_i, \mathbf{c}^E_j) / \tau)} \\
    &-\frac{1}{2}\sum_{i=1}^M \log \frac{\exp(s(\mathbf{c}^H_i, \mathbf{c}^E_i) / \tau)}{\sum_{j=1}^M \exp (s(\mathbf{c}^H_j, \mathbf{c}^E_i) / \tau)},
\end{aligned}
\end{equation}
where we use a similarity function based on L1 distance: $s(\mathbf{x}, \mathbf{y}) = -\|\mathbf{x} - \mathbf{y}\|_1$, and $\tau$ denotes the temperature hyperparameter. The final training objective for representation alignment is as follows:
\begin{equation}
    \mathcal{L}_\text{Align} = \mathcal{L}_\text{MSE} + \mathcal{L}_\text{CTR}.
\end{equation}

\paragraph{Two-stage Finetuning} 
In the zero-shot learning scenario, we adopt a two-stage finetuning strategy to facilitate the learning process. In the first stage, we only utilize the S2TT dataset $\mathcal{D}_\text{S2TT}$ to train the S2TT Model $\mathcal{F}$ and vocabulary adaptor $\mathcal{A}$ through the objectives $\mathcal{L}_\text{S2TT}$ and $\mathcal{L}_\text{CTC}$. In the second stage, we conduct training using both the S2TT and TTS datasets, with the following training objectives:
\begin{equation}
    \begin{aligned}
        &\mathcal{L}_\text{\name-ZS} = \mathbb{E}_{\mathcal{D}_\text{TTS}} \left[\mathcal{L}_\text{TTS}\right] \\
        &\quad \quad \quad +\mathbb{E}_{\mathcal{D}_\text{S2TT}} \left[\mathcal{L}_\text{S2TT} + \mathcal{L}_\text{CTC} + \mathcal{L}_\text{Align}\right].
    \end{aligned}   
\end{equation}
Throughout the entire training process, the S2ST data is not used. During inference, we can achieve zero-shot S2ST as illustrated in Figure \ref{fig:training}(c).

\section{Experiments}
\subsection{Datasets}

\begin{table}[t]
    \centering
\resizebox{\linewidth}{!}{
\begin{tabular}{c|c|c|c}
\toprule
\textbf{Direction} & \textbf{S2ST (src / tgt)} & \textbf{S2TT (src)} & \textbf{TTS (tgt)} \\
\midrule
\textbf{Fr$\rightarrow$En} & 264h / 174h & 264h & \multirow{3}{*}{191h} \\
\textbf{De$\rightarrow$En} & 184h / 112h & 184h &  \\
\textbf{Es$\rightarrow$En} & 113h / 70h  & 113h &  \\
\bottomrule
\end{tabular}}
    \caption{The speech hour statistics of all datasets. src: source speech; tgt: target speech.}
    \label{tab:data}
\end{table}

\begin{table*}[t]
\centering
\resizebox{\textwidth}{!}{
\begin{tabular}{l|l|ccc|cccc|c}
\toprule
\multirow{2}{*}{\textbf{ID}} & \multirow{2}{*}{\textbf{Model}} & \multicolumn{3}{c|}{\textbf{Training Data}} & \multicolumn{4}{c|}{\textbf{ASR-BLEU}} & \multirow{2}{*}{\textbf{Speedup}} \\
& & \textbf{S2ST} & \textbf{S2TT} & \textbf{TTS} & \textbf{Fr$\rightarrow$En} & \textbf{De$\rightarrow$En} & \textbf{Es$\rightarrow$En} & \textbf{Avg.} & \\
\midrule
\texttt{-} & Ground Truth & / & / & / & 84.52 & 75.53 & 88.54 & 82.86 & / \\
\midrule
\multicolumn{10}{c}{\textit{Supervised Learning}} \\
\midrule
\texttt{A1} & Translatotron 2~\citep{translatotron2} & $\checkmark$ & $\times$ & $\times$ & 26.12 & 16.92 & 22.98 & 22.01 & 1.00$\times$ \\
\texttt{A2} & UnitY~\citep{inaguma-etal-2023-unity} & $\checkmark$ & $\times$ & $\times$ & 26.90 & 16.36 & 24.06 & 22.44 & 0.98$\times$ \\
\texttt{A3} & S2UT~\citep{s2ut} & $\checkmark$ & $\times$ & $\times$ & 22.23 & 2.99 & 18.53 & 14.58 & 0.62$\times$ \\
\texttt{A4} & DASpeech~\citep{fang-etal-2023-daspeech} & $\checkmark$ & $\checkmark$ & $\checkmark^\dagger$ & 25.03 & / & / & / & \textbf{10.05$\times$} \\
\texttt{A5} & \textbf{\name} w/o pretrain & $\checkmark$ & $\times$ & $\times$ & 27.58** & 17.35** & 24.29** & 23.07 & 3.40$\times$\\
\texttt{A6} & \textbf{\name} w/ pretrain & $\checkmark$ & $\checkmark$ & $\checkmark^\dagger$ & \textbf{28.15**} & \textbf{18.16**} & \textbf{24.80**} & \textbf{23.70} & 3.40$\times$ \\

\midrule
\multicolumn{10}{c}{\textit{Zero-shot Learning}} \\
\midrule
\texttt{B1} & S2TT + G2P + TTS & $\times$ & $\checkmark$ & $\checkmark$ & 27.04 & 16.97 & 23.26 & 22.42 & \textbf{3.49$\times$} \\
\texttt{B2} & \textbf{\name-ZS} & $\times$ & $\checkmark$ & $\checkmark$ & \textbf{27.55**} & \textbf{17.67**} & \textbf{23.80**} & \textbf{23.01} & 3.40$\times$ \\
\bottomrule
\end{tabular}}
\caption{ASR-BLEU scores on CVSS Fr/De/Es$\rightarrow$En \texttt{test} set. $\dagger$: The TTS data used in the supervised learning scenario is different from that used in the zero-shot learning scenario as described in footnote~\ref{footnote:ttsdata}. ** means the improvements over the baseline system (\texttt{A1} in the supervised learning scenario and \texttt{B1} in the zero-shot learning scenario) are statistically significant ($p < 0.01$). }
\label{tab:main-results}
\end{table*}

The S2ST, S2TT, and TTS datasets utilized in our experiments are all sourced from the CVSS dataset~\citep{jia2022cvss}. CVSS is a large-scale S2ST dataset containing <source speech, target text, target speech> triples across 21 source languages to English. We conduct experiments on three language pairs: French$\rightarrow$English (Fr$\rightarrow$En), German$\rightarrow$English (De$\rightarrow$En), and Spanish$\rightarrow$English (Es$\rightarrow$En). Our experiments consist of two scenarios: \textit{supervised learning} and \textit{zero-shot learning} scenarios. The former involves training using S2ST data, while the latter involves training using only S2TT and TTS data.
\paragraph{S2ST Dataset} In the supervised learning scenario, we utilize triplet data from the CVSS Fr/De/Es$\rightarrow$En datasets to train the model.
\paragraph{S2TT Dataset} In the zero-shot learning scenario, we employ <source speech, target text> pairs from the CVSS Fr/De/Es$\rightarrow$En datasets as the S2TT data, with the target speech being discarded.
\paragraph{TTS Dataset} In the zero-shot learning scenario, we combine all <target text, target speech> pairs from CVSS X$\rightarrow$En (X$\notin$\{Fr, De, Es\}) datasets as the TTS data. We exclude the target speech-text pairs from CVSS Fr/De/Es$\rightarrow$En datasets, ensuring that there is no overlap in the target text between S2TT and TTS datasets. This guarantees that in the zero-shot learning scenario, the model has not implicitly utilized parallel <source speech, target speech> pairs for training. The statistical information for all datasets are presented in Table~\ref{tab:data}.


\subsection{Experimental Setup}

\paragraph{Data Processing}
For the source speech, we convert it to 16000Hz and compute 80-dimensional mel-filterbank features. For the target speech, we convert it to 22050Hz and transform the waveform into mel-spectrograms. Utterance-level and global-level cepstral mean-variance normalization are applied to the source speech and target speech, respectively. For the target text, the S2TT model employs a subword vocabulary of size 6k which is learned on the target text of CVSS. The TTS model employs a phoneme vocabulary of size 70. We follow~\citet{ren2021fastspeech2} to extract the duration, pitch, and energy information of the target speech.

\paragraph{Model Configuration}
The S2TT model comprises 12 Conformer encoder layers and 4 Transformer decoder layers. The TTS model follows the standard configuration of FastSpeech 2. The vocabulary adaptor contains 4 Transformer encoder layers and the upsample factor $\lambda$ is set to 5. The detailed hyperparameters can be found in Appendix~\ref{app:model_config}. The dropout is set to 0.3. The label smoothing of the S2TT model is set to 0.1. The HiFi-GAN~\citep{hifi-gan} vocoder pretrained on the VCTK dataset~\citep{Veaux2017CSTRVC} is used to generate waveform from the mel-spectrogram.

\paragraph{Training}
During the training process, the batch size for S2TT and S2ST data is set to 320k source audio frames, while the batch size for TTS data is set to 512. In the supervised learning scenario, the model is first pretrained on S2TT and TTS data\footnote{\label{footnote:ttsdata}It should be noted that in the supervised learning scenario, the TTS pretraining data are <target text, target speech> pairs sourced from S2ST data, distinct from the TTS data utilized in the zero-shot learning scenario.}, followed by finetuning using S2ST data. In the zero-shot learning scenario, a two-stage finetuning approach is employed as stated in Section \ref{sec:training}. During each training stage, the learning rate warms up to 1e-3 within 4k steps, and the training is halted if the validation loss does not decrease for 10 consecutive validations. We use Adam optimizer~\citep{adam} in all training stages. The temperature $\tau$ in contrastive learning is set to 0.1. All models are trained on 4 RTX 3090 GPUs.

\begin{table*}[t]
\centering
\small
\resizebox{\textwidth}{!}{
\begin{tabular}{l|ccc|ccc|ccc}
\toprule
\multirow{2}{*}{\textbf{Model}} & \multicolumn{3}{c|}{\textbf{Fr$\rightarrow$En}} & \multicolumn{3}{c|}{\textbf{De$\rightarrow$En}} & \multicolumn{3}{c}{\textbf{Es$\rightarrow$En}}  \\
& \textbf{S2TT} & \textbf{S2ST} & \textbf{$\Delta$}
& \textbf{S2TT} & \textbf{S2ST} & \textbf{$\Delta$}
& \textbf{S2TT} & \textbf{S2ST} & \textbf{$\Delta$} \\
\midrule
S2TT & 30.49 & / & / & 18.54 & / & / & 25.39 & / & / \\
\midrule
Translatotron 2 \citep{translatotron2} & 28.82 & 26.12 & 2.70 & \underline{18.66} & 16.92 & 1.74 & 25.82 & 22.98 & 2.84 \\
UnitY \citep{inaguma-etal-2023-unity} & \underline{30.37} & 26.90 & 3.47 & 17.95 & 16.36 & 1.59 & \textbf{26.59} & 24.06 & 2.53 \\
\textbf{\name} w/o pretrain & 29.98 & \underline{27.58} & \textbf{2.40} & 18.44 & \underline{17.35} & \textbf{1.09} & 25.73 & \underline{24.29} & \textbf{1.44} \\
\textbf{\name} w/ pretrain & \textbf{30.72} & \textbf{28.15} & \underline{2.57} & \textbf{19.41} & \textbf{18.16} & \underline{1.25} & \underline{26.51} & \textbf{24.80} & \underline{1.71} \\
\bottomrule
\end{tabular}}
\caption{BLEU scores of the S2TT results and ASR-BLEU scores of the S2ST results. The S2TT results come from the first pass of decoding. The best and second-best results are indicated by \textbf{bold} and \underline{underline}, respectively.}
\label{tab:gap}
\end{table*}

\paragraph{Evaluation}
We average the best 5 checkpoints based on validation loss for evaluation, and employ two evaluation metrics: ASR-BLEU and BLASER 2.0~\citep{communication2023seamlessm4t}. For ASR-BLEU, we first transcribes the translated speech into text using a pretrained ASR model\footnote{\url{https://dl.fbaipublicfiles.com/fairseq/wav2vec/wav2vec_vox_960h_pl.pt}}, and then calculates the BLEU score~\citep{papineni-etal-2002-bleu} and the statistical significance of translation results using the SacreBLEU toolkit\footnote{\url{https://github.com/mjpost/sacrebleu}}~\citep{post-2018-call}. For BLASER 2.0, we use the \texttt{blaser-2.0-ref} model\footnote{\url{https://huggingface.co/facebook/blaser-2.0-ref}} to evaluate the cross-lingual semantic similarity. The decoding speed is measured on the CVSS-C Fr$\rightarrow$En \texttt{test} set with a batch size of 1.

\begin{table}[t]
\centering
\resizebox{\linewidth}{!}{
\begin{tabular}{l|ccc}
\toprule
\multirow{2}{*}{\textbf{Model}} & \multicolumn{3}{c}{\textbf{BLASER 2.0}} \\
 & \textbf{Fr$\rightarrow$En} & \textbf{De$\rightarrow$En} & \textbf{Es$\rightarrow$En} \\
\midrule
\multicolumn{4}{c}{\textit{Supervised Learning}} \\
\midrule
Translatotron 2~\citep{translatotron2} & 3.1801 & 2.9034 & 3.2592 \\
UnitY~\citep{inaguma-etal-2023-unity} & 3.1749 & 2.8278 & 3.2310 \\
\textbf{ComSpeech} w/ pretrain & \textbf{3.1890} & \textbf{2.9281} & \textbf{3.2785} \\
\midrule
\multicolumn{4}{c}{\textit{Zero-shot Learning}} \\
\midrule
S2TT + G2P + TTS & \textbf{3.1716} & 2.8680 & 3.2172 \\
\textbf{ComSpeech-ZS} & 3.1681 & \textbf{2.9242} & \textbf{3.2555} \\
\bottomrule
\end{tabular}}
\caption{BLASER 2.0 scores on CVSS Fr/De/Es$\rightarrow$En \texttt{test} set.}
\label{tab:blaser}
\end{table}

\paragraph{Baseline Systems}
We primarily compare \name with two state-of-the-art two-pass S2ST models: Translatotron 2~\citep{translatotron2} and UnitY~\citep{inaguma-etal-2023-unity}. Their S2TT component is identical to the S2TT part of \name. Upon the S2TT model, a 2-layer Transformer encoder is employed to encode the S2TT decoder hidden states. Finally, UnitY utilizes a 2-layer decoder to generate the target units, while Translatotron 2 employs a 6-layer decoder to generate the target mel-spectrograms. Besides, we include the results of the single-pass model S2UT~\citep{s2ut} and the non-autoregressive two-pass model DASpeech~\citep{fang-etal-2023-daspeech} for comparison. We implement all above models with the open-source \texttt{fairseq}\footnote{\url{https://github.com/facebookresearch/fairseq}}~\citep{ott2019fairseq} library.

In the zero-shot learning scenario, we compare our proposed \name-ZS with a cascaded system: S2TT + G2P + TTS. Here, the S2TT and TTS models correspond to those used for initializing the \name-ZS model. The G2P is a model that converts English graphemes to phonemes. We achieve this conversion using the \texttt{g2p\_en}\footnote{\url{https://github.com/Kyubyong/g2p}} library.

\subsection{Main Results}

\paragraph{Results in the Supervised Learning Scenario}
Table \ref{tab:main-results} presents the ASR-BLEU scores on the CVSS \texttt{test} set. In the supervised learning scenario, it can be observed that: (1) \name outperforms Translatotron 2 and UnitY in translation quality across all three language pairs (\texttt{A1-A2} vs. \texttt{A5-A6}). Since the S2TT parts of these three models are the same, the performance improvement in S2ST is mainly attributed to the adoption of a more powerful TTS model for speech synthesis in \name. We also report the performance of S2TT and S2ST, along with their gap in Table \ref{tab:gap}. We observe that \name narrows the performance gap between S2TT and S2ST compared to previous models, further demonstrating the importance of incorporating a more powerful TTS for speech synthesis. (2) Benefiting from the parallel decoding capability of FastSpeech 2, \name achieves a 3.40$\times$ decoding speedup compared with Translatotron 2. (3) Compared to S2UT, \name shows significant improvements in translation quality (\texttt{A3} vs. \texttt{A5}), demonstrating the effectiveness of two-pass modeling in S2ST.  (4) Compared to DASpeech, which also employs FastSpeech 2 as the TTS module, \name shows a 3.1 ASR-BLEU improvement in translation quality (\texttt{A4} vs. \texttt{A6}). We believe the main reason is that DASpeech uses a phoneme vocabulary for the S2TT model to be compatible with FastSpeech 2's vocabulary, which may lead to a decrease in translation quality. Despite \name's slower decoding speed compared to DASpeech, we consider this is not the focus of our work. Theoretically, our proposed vocabulary adaptor can be also used to connect a more powerful non-autoregressive S2TT model with a TTS model. We leave this for future research.

\paragraph{Results in the Zero-shot Learning Scenario}
In the zero-shot learning scenario, we find that: (1) Despite not using any S2ST data, the translation quality of \name-ZS is comparable to that of \name in the supervised learning scenario, with only a 0.7 ASR-BLEU difference (\texttt{A6} vs. \texttt{B2}). (2) \name-ZS also surpasses the performance of Translatotron 2, UnitY, S2UT, and DASpeech (\texttt{A1-A4} vs. \texttt{B2}), further demonstrating the effectiveness of our proposed model architecture and training approach. (3) The translation quality of \name-ZS surpasses that of the cascaded system, possibly due to avoiding error accumulation. Meanwhile, \name-ZS is easier to deploy compared to the cascaded system, requiring only the deployment of a single model without the need to store intermediate results. 

\paragraph{Results of BLASER 2.0 Scores}
Table~\ref{tab:blaser} shows the BLASER 2.0 scores on the CVSS \texttt{test} set. We observe that our \name and \name-ZS consistently outperform other baseline systems in most cases. Since BLASER 2.0 evaluates directly based on speech rather than the transcribed text, this further validates that the speech generated by our model not only exhibits accurate translation but also reliable speech quality.

\begin{table}[]
    \centering
\resizebox{\linewidth}{!}{
\begin{tabular}{c|c|ccc}
\toprule
\textbf{MSE} & \textbf{CTR} ($s(\mathbf{x}, \mathbf{y})$) & \textbf{ASR-BLEU$\uparrow$} & \textbf{Alignment$\downarrow$} & \textbf{Uniformity$\downarrow$} \\
\midrule
$\times$ & $\times$ & 0.01 & 101.55 & \textbf{-56.75} \\
$\checkmark$ & $\times$ & 13.33 & 1.05 & -2.01 \\
$\times$     & $-\|\mathbf{x} - \mathbf{y}\|_1$ & 25.14 & 4.29 & -14.19 \\
$\times$     & $-\|\mathbf{x} - \mathbf{y}\|_2^2$ & 25.40 & 10.14 & -23.38 \\
$\times$     & $\mathbf{x}\cdot\mathbf{y}$ & 24.35 & 18.56 & -32.87 \\
$\checkmark$ & $-\|\mathbf{x} - \mathbf{y}\|_1$ & \textbf{27.55} & \textbf{0.85} & -3.97 \\
$\checkmark$ & $-\|\mathbf{x} - \mathbf{y}\|_2^2$ & 27.11 & 0.98 & -8.22 \\
$\checkmark$ & $\mathbf{x}\cdot\mathbf{y}$ & 27.13 & 0.92 & -9.30 \\
\bottomrule
\end{tabular}}
    \caption{Results on CVSS Fr$\rightarrow$En \texttt{test} set with different alignment training objectives.}
    \label{tab:objectives}
\end{table}

\subsection{Ablation Studies}
\paragraph{Training Objectives}

In the zero-shot learning scenario, achieving representation alignment is crucial for the final translation quality. In Table \ref{tab:objectives}, we explore different alignment training objectives. In addition to ASR-BLEU, we follow \citet{pmlr-v119-wang20k} to measure the alignment and uniformity of the representation space, defined as follows:
\begin{align}
    \text{Alignment} &= \mathbb{E}_{(\mathbf{c}^H_i, \mathbf{c}^E_i)} \left[\|\mathbf{c}^H_i - \mathbf{c}^E_i\|_1\right] \\
    \text{Uniformity} &= \log \mathbb{E}_{(\mathbf{c}^H_i, \mathbf{c}^E_j)} \left[e^{-\|\mathbf{c}^H_i - \mathbf{c}^E_j\|_1}\right]
\end{align}
From the results in Table \ref{tab:objectives}, it can be observed that using only MSE loss results in a relatively low ASR-BLEU. Using only contrastive learning also leads to a performance drop of around 2 ASR-BLEU, as the alignment score is high in this case. Hence, both contrastive learning and MSE loss are indispensable. Specifically, our experiments reveal that using L1 distance as the similarity function yields better results than L2 distance and dot product. In this case, even though the uniformity score is higher, the alignment score is the lowest. We speculate that the alignment of representation space plays a more crucial role in the ultimate zero-shot transfer performance.

\paragraph{Pretraining and Finetuning Strategies}
\begin{table}[]
    \centering
\resizebox{\linewidth}{!}{
\begin{tabular}{cc|c|cc}
\toprule
\multicolumn{2}{c|}{\textbf{Pretrain}} & \multirow{2}{*}{\textbf{\makecell{Two-stage \\Finetune}}} & \multicolumn{2}{c}{\textbf{ASR-BLEU}} \\
\textbf{S2TT} & \textbf{TTS} & & \textbf{\name} & \textbf{\name-ZS} \\
\midrule
$\times$ & $\times$ & $\times$ & 27.58 & 23.43 \\
$\times$ & $\checkmark$ & $\times$ & 27.43 &  0.19 \\
$\checkmark$ & $\times$ & $\times$ & 28.03 & 25.67 \\
$\checkmark$ & $\checkmark$ & $\times$ & \textbf{28.15} & 27.10 \\
$\checkmark$ & $\checkmark$ & $\checkmark$ & / & \textbf{27.55} \\
\bottomrule
\end{tabular}}
    \caption{ASR-BLEU scores on CVSS Fr$\rightarrow$En \texttt{test} set with different pretraining and finetuning strategies.}
    \label{tab:pretrain}
\end{table}

ComSpeech supports S2TT and TTS pretraining. We explore the impact of pretraining in both supervised and zero-shot learning scenarios. As shown in Table \ref{tab:pretrain}, using only TTS pretraining often has a negative effect, especially in the zero-shot learning scenario. On the other hand, S2TT pretraining brings a significant performance improvement, and combining both S2TT and TTS pretraining results in additional performance gains. Overall, pretraining has a greater impact on performance in the zero-shot learning scenario. Additionally, our proposed two-stage fine-tuning strategy leads to a performance improvement of 0.45 ASR-BLEU in the zero-shot learning scenario.



\subsection{Effects of the Size of S2TT and TTS Data}

In this section, we explore the effects of the size of S2TT and TTS data on \name-ZS and the cascaded system (S2TT + G2P + TTS).

\paragraph{S2TT Data Size}

\begin{figure}[t]
    \centering
    \includegraphics[width=\linewidth]{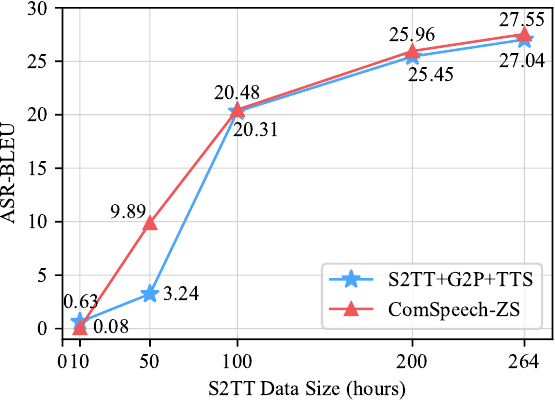}
    \caption{ASR-BLEU scores on CVSS Fr$\rightarrow$En \texttt{test} set with different amounts of S2TT data.}
    \label{fig:st_datasize}
\end{figure}

As shown in Figure \ref{fig:st_datasize}, the size of S2TT data significantly impacts the performance of both the cascaded system and \name-ZS. With only 10 hours of S2TT data, both systems have poor translation quality. However, as the amount of S2TT data increases, \name-ZS consistently achieves higher translation quality than the cascaded system, demonstrating the advantages of end-to-end modeling.

\begin{figure}[t]
    \centering
    \includegraphics[width=\linewidth]{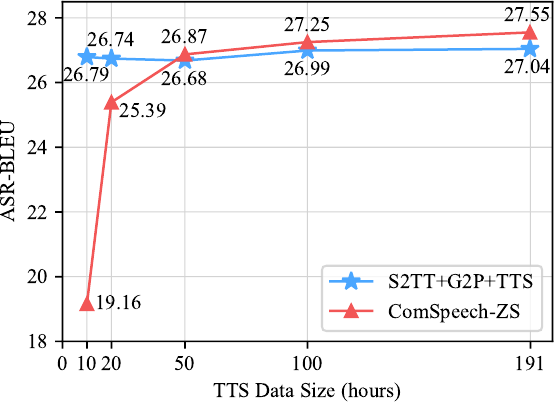}
    \caption{ASR-BLEU scores on CVSS Fr$\rightarrow$En \texttt{test} set with different amounts of TTS data.}
    \label{fig:datasize}
\end{figure}

\paragraph{TTS Data Size}
As shown in Figure \ref{fig:datasize}, the performance of the cascaded system remains relatively stable with the expansion of TTS data size, while \name-ZS gradually improves with the increase in TTS data. When TTS data is limited, \name-ZS performs worse than the cascaded system, and we speculate this may be due to the imbalance in the S2TT and TTS data scales during finetuning. As the TTS data exceeds 50 hours, \name-ZS outperforms the cascaded system, and the gap between them grows as the TTS data size increases. This indicates that \name-ZS has a higher upper limit than the cascaded system.

\section{Related Work}

\paragraph{Speech-to-Speech Translation} S2ST extends S2TT~\citep{fang-etal-2022-STEMM, fang-and-feng-2023-back, fang-and-feng-2023-understanding, zhou-etal-2023-cmot} which further generates the target speech. \citet{translatotron} first proposes direct S2ST with a sequence-to-sequence model. \citet{uwspeech, s2ut, lee-etal-2022-textless} propose using the discrete representation of speech as the prediction target and achieve better performance. \citet{huang2023chch, zhu-etal-2023-diffs2ut, wu-2023-duplex, fang-etal-2023-daspeech, fang-etal-2024-ctc} adopt non-autoregressive models or diffusion models to generate the target speech for faster decoding speed. To make training easier, \citet{translatotron2, inaguma-etal-2023-unity} introduce two-pass S2ST models that generate target text and target speech successively. 
To further enhance S2ST, researchers introduce techniques like pretraining \citep{speech2s, vallex, polyvoice} and data augmentation \citep{DBLP:conf/interspeech/PopuriCWPAGHL22, DBLP:conf/interspeech/JiaDBC0CM22, DBLP:conf/interspeech/DongYKWBZ22, nguyen2022improving, communication2023seamlessm4t} to alleviate the data scarcity.
\citet{zhang-etal-2024-streamspeech, ma-etal-2024-a} achieve simultaneous speech-to-speech translation with multi-task learning and non-autoregressive streaming Transformer, respectively. Our work extends the research on two-pass S2ST, introducing a more general model architecture and a novel training method that achieves zero-shot S2ST without parallel speech data.

\paragraph{Zero-shot Speech Translation}
\citet{DBLP:journals/corr/abs-2107-06010, wang-etal-2022-discrete, duquenne-etal-2022-modules, DBLP:journals/corr/abs-2310-03724} explore cross-modal alignment between speech and text, achieving zero-shot S2TT using only ASR and MT data. For S2ST, \citet{diwan2023unitbased, nachmani2024translatotron} focus on achieving zero-shot S2ST using only monolingual speech data in both source and target languages. To the best of our knowledge, we are the first to study achieving zero-shot S2ST using only S2TT and TTS data.

\section{Conclusion}
In this paper, we first introduce a novel S2ST model architecture, named \name, which incorporates a CTC-based vocabulary adaptor capable of connecting arbitrary S2TT and TTS models to obtain an S2ST model. Furthermore, we propose a training method using only S2TT and TTS data. By aligning in the TTS encoder's representation space using contrastive learning, we achieve zero-shot S2ST without relying on parallel speech data. Experimental results on the CVSS dataset demonstrate that \name surpasses previous S2ST models in both translation quality and decoding efficiency. Moreover, in the zero-shot learning scenario, our model achieves performance close to the supervised learning scenario and surpasses the cascaded system. 
In summary, we hope that our method can inspire future work in direct S2ST in terms of model architecture and training approach, making more comprehensive use of existing S2TT and TTS pretrained models and training data to improve S2ST. 
In the future, we will explore building direct S2ST models based on more powerful S2TT and TTS models.

\section*{Limitations}
Although our model achieves satisfactory performance in both supervised learning and zero-shot learning scenarios, there are still some limitations: (1) In cases where TTS data is limited, the performance of \name-ZS still lags behind the cascaded system. In the future, we will explore how to improve performance by balancing the S2TT and TTS training data. (2) Currently, our method cannot preserve the speaker characteristics of the source speech. We will explore this in the future.

\section*{Acknowledgement}
We thank all the anonymous reviewers for their insightful and valuable comments. This paper is supported by National Natural Science Foundation of China (Grant No.62376260).

\bibliography{custom}
\newpage
\appendix


\onecolumn

\section{Effects of the Hyperparameters of Vocabulary Adaptor}
The vocabulary adaptor has two important hyperparameters: the number of Transformer encoder layers $L$ and the upsampling factor $\lambda$ for the input sequence. We investigate the influence of these hyperparameters in the supervised learning scenario without S2TT and TTS pretraining. According to the performance on the CVSS Fr$\rightarrow$En \texttt{dev} set as shown in Table~\ref{tab:adaptor}, we choose $L=4$ and $\lambda=5$ in our expriments.

\begin{table}[h] 
\centering
\begin{tabular}{cc|cc}
\toprule
\textbf{$L$} & \textbf{ASR-BLEU} & \textbf{$\lambda$} & \textbf{ASR-BLEU} \\
\midrule
2 & 27.58 & 2 & 11.94 \\
4 & \textbf{28.04} & 5 & \textbf{28.04} \\
6 & 28.02 & 8 & 27.75 \\
\bottomrule
\end{tabular}
\caption{Results on CVSS Fr$\rightarrow$En \texttt{dev} set with different hyperparameters of the vocabulary adaptor.}
\label{tab:adaptor}
\end{table}


\section{Hyperparameters}
\label{app:model_config}

We list the hyperparameters of \name and other baseline models in Table~\ref{tab:hyperparameter}.

\begin{table*}[h]
\centering
\resizebox{\textwidth}{!}{
\begin{tabular}{c|c|ccccc}
\toprule
\multicolumn{2}{c|}{\textbf{Hyperparameters}} & \textbf{S2UT} & \textbf{UnitY} & \textbf{Translatotron 2} & \textbf{DASpeech} & \textbf{ComSpeech} \\
\bottomrule
\multirow{8}{*}{S2TT Encoder} & conv\_kernel\_sizes & (5, 5) & (5, 5) & (5, 5) & (5, 5) & (5, 5) \\
 & encoder\_type & conformer & conformer & conformer & conformer & conformer \\
 & encoder\_layers & 12 & 12 & 12 & 12 & 12 \\
 & encoder\_embed\_dim & 256 & 256 & 256 & 256 & 256 \\
 & encoder\_ffn\_embed\_dim & 2048 & 2048 & 2048 & 2048 & 2048 \\
 & encoder\_attention\_heads & 4 & 4 & 4 & 4 & 4 \\
 & encoder\_pos\_enc\_type & relative & relative & relative & relative & relative \\
 & depthwise\_conv\_kernel\_size & 31 & 31 & 31 & 31 & 31 \\
\midrule
\multirow{6}{*}{S2TT Decoder} & decoder\_layers & 4 & 4 & 4 & 4 & 4 \\
 & decoder\_embed\_dim & 512 & 512 & 512 & 512 & 512 \\
 & decoder\_ffn\_embed\_dim & 2048 & 2048 & 2048 & 2048 & 2048 \\
 & decoder\_attention\_heads & 8 & 8 & 8 & 8 & 8\\
 & label\_smoothing & 0.1 & 0.1 & 0.1 & 0.0 & 0.1 \\
 & s2t\_loss\_weight & 8.0 & 8.0 & 0.1 & 1.0 & 1.0 \\
\midrule
\multirow{4}{*}{Vocabulary Adaptor} & encoder\_layers & - & - & - & - & 4 \\
& encoder\_embed\_dim & - & - & - & - & 512 \\
& encoder\_ffn\_embed\_dim & - & - & - & - & 2048 \\
& encoder\_attention\_heads & - & - & - & - & 8 \\
\midrule
\multirow{4}{*}{TTS Encoder} & encoder\_layers & - & 2 & 2 & 4 & 4 \\
 & encoder\_embed\_dim & - & 512 & 512 & 256 & 256 \\
 & encoder\_ffn\_embed\_dim & - & 2048 & 2048 & 1024 & 1024 \\
 & encoder\_attention\_heads & - & 8 & 8 & 4 & 4 \\
\midrule
\multirow{11}{*}{TTS Decoder} & decoder\_layers & 6 & 2 & 6 & 4 & 4 \\
 & decoder\_embed\_dim & 512 & 512 & 512 & 256 & 256 \\
 & decoder\_ffn\_embed\_dim & 2048 & 2048 & 2048 & 1024 & 1024 \\
 & decoder\_attention\_heads & 8 & 8 & 8 & 4 & 4 \\
 & label\_smoothing & 0.1 & 0.1 & - & - & - \\
 & n\_frames\_per\_step & 1 & 1 & 5 & 1 & 1 \\
 & unit\_dictionary\_size & 1000 & 1000 & - & - & - \\
 & var\_pred\_hidden\_dim & - & - & - & 256 & 256 \\
 & var\_pred\_kernel\_size & - & - & - & 3 & 3 \\
 & var\_pred\_dropout & - & - & - & 0.5 & 0.5 \\
 & s2s\_loss\_weight & 1.0 & 1.0 & 1.0 & 5.0 & 1.0 \\
\bottomrule
\end{tabular}}
\caption{Hyperparameters of ComSpeech and baseline models.}
\label{tab:hyperparameter}
\end{table*}

\end{document}